\documentclass[letterpaper, 10 pt, conference]{ieeeconf}
\usepackage{url,amsmath,amsfonts,amssymb,bm, mathtools, algorithm,algpseudocode,mathrsfs,setspace,bbm,tikz,forest}
\IEEEoverridecommandlockouts

\newtheorem{theorem}{Theorem}
\newtheorem{lemma}{Lemma}

\newtheorem{assumption}{Assumption}

\newcommand{\state}{\mathcal{S}}
\newcommand{\M}{\mathcal{M}}

\newcommand{\E}{{\mathbb E}}
\newcommand{\X}{{\mathcal X}}

\renewcommand{\P}{{\mathcal P}}

\newcommand{\A}{\mathcal{A}}

\newcommand{\rv}[1]{\bm{\mathrm{#1}}}

\begin{document}
\onecolumn

\title{A Sample-Efficient Algorithm for Episodic Finite-Horizon MDP with Constraints}
\author{\normalsize Krishna C. Kalagarla, Rahul Jain,  Pierluigi Nuzzo\\
Ming Hsieh Department of Electrical and Computer Engineering, University of Southern California, Los Angeles \\
Email: \{kalagarl,rahul.jain,nuzzo\}@usc.edu}
\maketitle

\begin{abstract}
Constrained Markov Decision Processes (CMDPs) formalize sequential decision-making problems whose objective is to minimize a cost function while satisfying constraints on various cost functions. In this paper, we consider the setting of episodic fixed-horizon CMDPs. We propose an online algorithm which leverages the linear programming formulation of finite-horizon CMDP for repeated optimistic planning to provide a probably approximately correct (PAC) guarantee on the number of episodes needed to ensure an $\epsilon$-optimal policy,  i.e., with resulting objective value within $\epsilon$ of the optimal value and satisfying the constraints within $\epsilon$-tolerance, with probability at least $1-\delta$. The number of episodes needed is shown to be of the order $\tilde{\mathcal{O}}\big(\frac{|\state||\A|C^{2}H^{2}}{\epsilon^{2}}\log\frac{1}{\delta}\big)$, where $C$ is the upper bound on the number of possible successor states for a state-action pair. Therefore, if $C \ll |S|$, the number of episodes needed have a linear dependence on the state and action space sizes $|S|$ and $|A|$, respectively, and quadratic dependence on the time horizon $H$.
\end{abstract}

\section{Introduction}

Markov decision processes (MDPs)~\cite{Puterman:1994:MDP:528623} offer a natural framework to express sequential decision-making problems and reason about autonomous system behaviors. However, the single cost objective of a traditional MDP formulation may fall short of fully capturing problems with multiple conflicting objectives and additional constraints that must be satisfied. Consider, for example, an autonomous car that is required to reach a destination at the earliest, but also satisfy a set of safety requirements and fuel consumption constraints, while keeping a desired comfort level~\cite{pmlr-v97-le19a}. The framework of constrained MDPs (CMDPs)~\cite{altman1999constrained} extended MDPs by considering additional constraints on the expected long-term performance of a policy. The objective in a CMDP is to minimize the expected cumulative cost while satisfying the additional constraints. In this paper, we consider episodic finite-horizon CMDPs, where an agent interacts with a CMDP repeatedly in episodes of fixed length, a setting that can model a large number of repetitive tasks such as goods delivery or customer service. 

We address the problem of online learning of CMDPs with unknown transition probabilities, by requiring only observed trajectories 
rather than sampling the transition function for any state-action pair from a generative model, which may not always be available. 
An important question which arises in online learning is the exploration-exploitation dilemma, i.e., the trade-off between exploration, to gain more information about the model, and exploitation, to minimize the cost. In this respect, the performance of learning algorithms is commonly evaluated in terms of (i) regret, i.e., the difference between the cumulative cost of the agent and that of the optimal policy in hindsight, and (ii) sample complexity, i.e., the number of steps for which the learning agent may not play a near-optimal policy. We consider a policy to be near optimal if the expected cumulative cost is close to the optimal and the constraints are satisfied within a small tolerance. 
In this paper, we address sample-efficiency by proposing an algorithm that provide Probably Approximately Correct (PAC) guarantees. 

Our algorithm leverages the concept of \textit{optimism-in-the-face-of-uncertainty} \cite{lai1985asymptotically,auer2009near} to balance exploration and exploitation. The learning agent repeatedly defines a set of statistically plausible transition models given the observations made so far.  It then chooses an optimistic transition probability model and optimistic policy with respect to the given constrained MDP problem. This planning step is formulated as a linear programming (LP) problem in occupancy measures, whose solution gives the desired optimistic policy. This policy is then executed for multiple episodes until a state-action pair has been visited sufficiently often. The total visitation counts are then updated 
and these steps are repeated.

We show that the number of episodes in which the learning agent plays an $\epsilon$-suboptimal policy is upper bounded by $\tilde{\mathcal{O}}\big(\frac{|\state||\A|C^{2}H^{2}}{\epsilon^{2}}\log\frac{1}{\delta}\big) $ with probability at least $1-\delta$, where $C$ is the upper bound on the number of possible successor states for a state-action pair.

\textbf{Contribution.} In this paper, we present one of the first online algorithms with PAC guarantees for episodic constrained MDPs with unknown transition probabilities. We build on the work of \cite{dann2015sample} which provides a PAC algorithm for unconstrained episodic MDPs.
However, differently from planning based on the Bellman optimality equations~\cite{dann2015sample}, we address the presence of constraints by  formulating an optimistic planning problem as an LP in occupancy measures. Consequently, our formulation leverages a novel 
construction for the set of plausible transition models and results in a
sample complexity that is quadratic in the time-horizon $H$, thus improving 
on the cubic 
bounds previously obtained with regret-based formulations (e.g., see~\cite{efroni2020exploration}).  

\textbf{Related Work.}
There has been significant work on efficient learning for unconstrained MDPs. Algorithms like UCBVI~\cite{10.5555/3305381.3305409}, UBEV~\cite{dann2017unifying}, EULER~\cite{pmlr-v97-zanette19a} and EULER-GP~\cite{efroni2019tight} focus on the setting of regret analysis for unconstrained episodic finite-horizon MDPs.  The setting of PAC algorithms for unconstrained MDPs is addressed by~\cite{dann2015sample,brafman2002r,strehl2009reinforcement}. While these previously mentioned algorithms are model-based RL algorithms, model-free algorithms UCB-H and UCB-B~\cite{jin2018q} have also been shown to be sample efficient.

Sample efficient exploration in CMDPs has recently started to receive attention. The regret analysis for multiple model-based and model-free algorithms \cite{efroni2020exploration} has been performed in the setting of episodic CMDPs with stochastic cost functions and unknown transition probabilities. Our work addresses PAC complexity, and is therefore  complementary to \cite{efroni2020exploration}. There has also been other parallel works on regret analysis for constrained MDPs in the setting of average cost \cite{singh2020learning}, adversarial cost with tabular MDPs \cite{qiu2020upper} and adversarial cost with linear MDPs~\cite{ding2020provably}.

There has also been work on constrained MDPs with stronger requirements. Algorithm C-UCRL \cite{Zheng2020ConstrainedUC} has been shown to have sublinear regret and satisfy the constraints even while learning, albeit in the setting of known transition probabilities and unknown cost functions. Regret optimal algorithm for constrained MDPs with concave objectives and convex  and hard constraints (knapsacks) (i.e., problems with a fixed budget such that the learning is stopped as soon as the budget is consumed) is studied by \cite{brantley2020constrained}.
Several of these regret algorithms can be modified following an idea from \cite{jin2018q} to provide PAC guarantees for constrained MDP with time-horizon dependence of at least $H^3$. But, this procedure is impractical as it entails saving an extremely large number of policies and uniformly sampling them to get the PAC optimal policy. 

There are also policy optimization and Lagrangian based works on constrained MDPs \cite{borkar2005actor,achiam2017constrained, tessler2018reward,miryoosefi2019reinforcement} but these lack regret or PAC analysis.

\section{Preliminaries}
In this section, we introduce preliminary concepts from finite-horizon MDPs and CMDPs. 


\subsection{Notation}
We denote the set of natural numbers by $\mathbb{N}$ and use $h \in \left[1:H\right]$ and $k \in \mathbb{N}$ to denote time-step inside an episode and phase index respectively. The indicator function $\mathbb{I}(s = s_1)$ is $1$ when $s = s_1$ and 0 otherwise. The probability simplex over set $S$ is denoted by $\Delta_{S}$. We use the notation $\tilde{\mathcal{O}} $ which is similar to the usual $\mathcal{O} $ notation but ignores logarithmic factors.

\subsection{Finite-Horizon MDPs} We consider episodic finite-horizon MDPs~\cite{Puterman:1994:MDP:528623}, which can be formally defined by a tuple $\M = (\state,\A,H,s_{1},p,c)$, where $\state$ and $\A$ denote the finite state and action spaces, respectively. The agent interacts with the environment in episodes of length $H$, with each episode starting with the same initial state $s_{1}$. The non-stationary transition probability is denoted by $p$ where $p_{h}(s'|s,a)$ is the the probability of transitioning to state $s'$ upon taking action $a$ at state $s$ at time step $h$. Further, we denote by $Succ(s,a)$ the set of possible successor states of state $s$ and action $a$. The maximum number of possible successor states is denoted by $C = \max_{s,a}|Succ(s,a)|$. The non-stationary cost of taking action $a$ in state $s$ at time step $h \in \left[ 1: H \right]$ 
is a random variable $C_{h}(s,a) \in \left[ 0, 1 \right] $, with mean $c_{h}(s,a)$. Finally, we set 
$c =  c_1,\ldots,c_H $. 

A non-stationary randomized policy $\pi = (\pi_{1}, \ldots , \pi_{H}) \in \Pi$ where $\pi_{i} : \state \to \Delta_{\A}$, maps each state to a probability simplex over the action space. We denote by $a_{h} \sim \pi_{h}(s_h)$, the action taken at time step $h$ at state $s_h$ according to policy $\pi$. For a state $s \in \state$ and time step $h \in \left[ 1 : H \right]$, the value function of a non-stationary randomized policy, $V_{h}^{\pi}(s;c,p)$  (when clear, $\pi,c,p$ is omitted) is defined as:
$$V_{h}^{\pi}(s;c,p) = \E \left[\sum_{i=h}^{H} c_{i}(s_{i},a_{i}) | s_h = s,\pi,p \right], $$
where the expectation is over the environment and policy randomness. Similarly, for a state $s \in \state$, an action $a \in \A$ and time step $h \in \left[ 1 : H \right]$, the Q-value function is defined as $Q_{h}^{\pi}(s,a;c,p) = $:
$$ c_{h}(s,a) + \E \left[\sum_{i=h+1}^{H} c_{i}(s_{i},a_{i}) | s_h = s,a_h = a,\pi,p \right]. $$ 
There always exists an optimal non-stationary deterministic policy $\pi^{*}$ \cite{Puterman:1994:MDP:528623} such that $V_{h}^{\pi^{*}}(s) = V_{h}^{*}(s) = \text{inf}_{\pi} V_{h}^{\pi}(s)$ and 
$Q_{h}^{\pi^{*}}(s,a) = Q_{h}^{*}(s,a) = \text{inf}_{\pi} Q_{h}^{\pi}(s,a)$.
The Bellman optimality equations \cite{Puterman:1994:MDP:528623} enable us to compute the optimal policy by backward induction:
 \begin{align*}
      &V_{h}^{*}(s) = \text{min}_{a \in \A} \left[c_{h}(s,a) + p_{h}(\cdot|s,a)V_{h+1}^{*}\right],\\
      &Q_{h}^{*}(s,a) =  c_{h}(s,a) + p_{h}(\cdot|s,a)V_{h+1}^{*},
 \end{align*}
 where $V_{H+1}^{*}(s) = 0$ and $V_{h}^{*}(s) = \text{min}_{a \in \A}Q_{h}^{*}(s,a)$. The optimal policy $\pi^{*}$ is thus greedy with respect to $Q_{h}^{*}$.
 
 \subsection{Finite-Horizon Constrained MDPs}
 A finite-horizon constrained MDP is a finite-horizon MDP along with additional $I$ constraints \cite{altman1999constrained} expressed by pairs of constraint cost functions and thresholds, $\{d_{i},l_{i}\}_{i=1}^{I}$. The cost of taking action $a$ in state $s$ at time step $h \in \left[ 1: H \right]$ with respect to the $i^{th}$ constraint cost function is a random variable $D_{i,h}(s,a) \in \left[ 0, 1 \right] $, with mean $d_{i,h}(s,a)$. The total expected cost of an episode under policy $\pi$ with respect to cost functions $c,d_i, i \in \left[ 1:I \right] $ is the respective value function from the initial state $s_1$,  i.e., $V_{1}^{\pi}(s_1;c), V_{1}^{\pi}(s_1;d_i), i \in \left[ 1: I \right] $ respectively (by definition). The objective of a CMDP is to find a policy which minimizes the total expected objective cost under the constraint that the total expected constraint costs are below the respective desired thresholds. Formally,
\begin{equation}\label{eq:obj}
    \begin{aligned}
   \pi^{*} \in \underset{\pi \in \Pi }{\text{ argmin }} \quad & V_{1}^{\pi}(s_{1};c,p)\\ \textrm{s.t.} \quad  & V_{1}^{\pi}(s_{1};d_{i},p) \leq l_{i} \quad \forall i \in \left[ 1: I \right].
\end{aligned}
\end{equation}
The optimal value is $V^{*} = V_{1}^{\pi^{*}}(s_{1};c,p) $. The optimal policy may be randomized \cite{altman1999constrained}, i.e., an optimal deterministic policy may not exist as in the case of the finite-horizon MDP. Further, the Bellman optimality equations do not hold due to the constraints. Thus, we cannot leverage backward induction as before to find an optimal policy. A linear programming approach has been shown \cite{altman1999constrained} to find an optimal policy.

\paragraph*{Linear Programming for CMDPs}
Occupancy measures \cite{altman1999constrained} allow formulating the optimization problem \eqref{eq:obj} as a linear program (LP). Occupancy measure $q^{\pi}$ of a policy $\pi$ in a finite-horizon MDP is defined as the expected  number of visits to a state-action  pair $(s,a)$ in an episode at time step $h$. Formally,
\begin{align*}
q^{\pi}_{h}(s,a;p) &= \E \left[  \mathbb{I}\{s_h = s, a_h =a \} | s_1=s_1,\pi,p \right] = Pr\left[s_h= s,a_h=a|s_1 = s_1,\pi,p \right].    
\end{align*} 
It is easy to see that the occupancy measure $q^{\pi}$ of a policy $\pi$ satisfy the following properties expressing non-negativity and flow conservation respectively:
\begin{align*}
    q^{\pi}_{h}(s,a) &\geq 0, \quad \forall (s,a,h) \in \state \times \A \times \left[1: H \right], \\
    q^{\pi}_{1}(s,a) &= \pi_{1}(a|s)\mathbb{I}(s = s_1), \quad \forall (s,a) \in \state \times \A ,\\
    \sum_{a}q^{\pi}_{h}(s,a) &= \sum_{s',a'}p_{h-1}(s|s'a')q^{\pi}_{h-1}(s',a'),
    \quad \forall s \in \state,h \in \left[ 2:H \right],
\end{align*}
where $\mathbb{I}(s = s_1)$ is the initial state distribution. The space of the occupancy measures satisfying the above constraints is denoted by $\Delta(\M)$. A policy $\pi$ generates an occupancy measure $q \in \Delta(\M)$ if  
\begin{equation}\label{eq:occu}
\pi_h(a|s) = \frac{q_h(s,a)}{\sum_{b}q_h(s,b)}, \quad \forall (s,a,h) \in \state \times \A \times \left[1: H \right].
\end{equation}
Thus, there exists a unique generating policy for all occupancy measures in $\Delta(\M)$ and vice versa. Further, the total expected cost of an episode under policy $\pi$ with respect to cost function $c$ can be expressed in terms of the occupancy measure as follows:
$$V_{1}^{\pi}(s_{1};c,p) = \sum_{h,s,a}q^{\pi}_{h}(s,a;p)c_h(s,a). $$
The optimization problem \eqref{eq:obj} can then be reformulated as a linear program \cite{altman1999constrained,zimin2013online} as follows:
\begin{align*}
 q^{*} \in & \underset{q \in \Delta(\M) }{\text{ argmin }} \quad    \sum_{h,s,a}q_{h}(s,a)c_h(s,a),   \\
 \textrm{s.t.} &\sum_{h,s,a}q_{h}(s,a)d_{i,h}(s,a) \leq l_i \quad \forall i \in \left[ 1: I \right].
\end{align*}
The optimal policy $\pi^{*}$ can be obtained from $q^{*}$ following \eqref{eq:occu}. 

\section{The Learning Problem} 
 
We consider the setting where an agent repeatedly interacts with a finite-horizon CMDP $\M = (\state,\A,H,s_1,p,c,\{d_{i},l_{i}\}_{i=1}^{I})$ with stationary transition probability (i.e., $p_h = p, \forall h \in \left[1:H\right]$) in episodes of fixed length $H$, starting from the same initial state $s_1$. For simplicity of analysis,\footnote{The complexity of learning the transition probability dominates the complexity of learning the cost functions \cite{auer2005online}. The algorithm can be readily extended to the setting of unknown cost functions by using an optimistic lower bound of the cost function obtained from its empirical estimate in place of the known cost function. 
} we assume that the cost functions $c,\{d_{i}\}_{i=1}^{I} $ are known to the learning agent, but the transition probability $p$ is unknown. The agent estimates the transition probability in an online manner by observing the trajectories over multiple episodes. 

The main objective is to design an online learning algorithm such that for given $\epsilon,\delta \in (0,1)$ and CMDP $\M$, the number of episodes for which the agent follows an $\epsilon$-suboptimal policy is bounded above by a polynomial (up to logarithmic factors) in the relevant quantities $(|\state|,|\A|,H,\frac{1}{\epsilon},\frac{1}{\delta})$  with high probability, i.e., with probability at least $1-\delta$ (PAC guarantee). A policy $\pi$ is said to be $\epsilon$-optimal if the total expected objective cost of an episode under policy $\pi$ is within $\epsilon$ of the optimal value, i.e., $V_{1}^{\pi}(s_1;c,p) \leq V^{*} + \epsilon$ and the constraints are satisfied within an $\epsilon$ tolerance, i.e., $V_{1}^{\pi}(s_1;d_i,p) \leq l_i + \epsilon, \forall \ i \in \left[1:I \right]$. We make the following assumption of feasibility. 

\begin{assumption}\label{asm}
The given CMDP $\M$ is feasible, i.e., there exists a policy $\pi$ such that the constraints are satisfied. 
\end{assumption}

\section{The {\tt UC-CFH} Algorithm}
 
\paragraph*{Algorithm Description.}
 
 We consider an adaptation of the model-based algorithm UCFH \cite{dann2015sample} to the setting of CMDP, which we call Upper-Confidence Constrained Fixed-Horizon episodic reinforcement learning ({\tt UC-CFH}) algorithm. The algorithm leverages the approach of \textit{optimism-in-the-face-of-uncertainty} \cite{auer2009near} to balance exploration and exploitation. 
 
 The algorithm operates in phases indexed by $k$ and whose length is not fixed, but instead depends on the observations made until the current episode. 
 Each phase consists of three stages: planning, policy execution, and update of the visitation counts. 
 
For each phase $k$, {\tt UC-CFH} defines a set of plausible transition models based on the number of visits to state-action pairs $(s,a)$ and transition tuples $(s,a,s')$ so far. A policy $\pi^{k}$ is then chosen by solving an optimistic planning problem, which is expressed as an LP problem (lines 13-16 in Algorithm 1). The planning problem, referred to as  \textit{ConstrainedExtendedLP} in the algorithm, is detailed below.  
 
 The algorithm maintains two types of visitation counts. Counts $v(s,a)$ and  $v(s,a,s')$ 
 are the number of visits to state-action pairs $(s,a)$ and transition tuples $(s,a,s')$, respectively, since the last update of state-action pair $(s,a)$. Counts $n(s,a)$ and $n(s,a,s')$ are the total number of  visits to state-action pairs $(s,a)$ and transition tuples $(s,a,s')$, respectively, before the update of state-action pair $(s,a)$. These visitation counts are all initialized to zero.
 
 During the policy execution stage of phase $k$ (lines 18-27 in Algorithm 1), the agent executes the current policy $\pi^{k}$, observes the tuples $(s_t,a_t,s_{t+1})$, and updates the respective visitation counts $v(s_t,a_t)$ and $v(s_t,a_t,s_{t+1})$. This policy $\pi^{k}$ is executed until a state action pair $(s,a)$ has been visited often enough since the last update of $(s,a)$, i.e., $v(s,a)$ is large enough (lines 26-27 in Algorithm 1). 
 
 In the next stage of phase $k$ (lines 29-33 in Algorithm 1), the visitation counts $n(s,a), n(s,a,s')$ corresponding to the sufficiently visited state action pair $(s,a)$ are updated as $n(s,a) = n(s,a) + v(s,a)$,  $n(s,a,s') = n(s,a,s') + v(s,a,s')$ and visitation counts $v(s,a),v(s,a,s') $ are reset to 0. This iteration of
planning-execution-update describes a phase of the algorithm.

\paragraph*{Optimistic Planning.} 
At the start of each phase $k$, {\tt UC-CFH} estimates the true transition model by its empirical average as:
$$\bar{p}^{k}(s'|s,a) = \frac{n^{k}(s,a,s')}{\max\{1,n^{k}(s,a)\}}, \quad  \forall (s,a,s') \in \state \times \A \times \state. $$
The algorithm further defines confidence intervals for the transition probabilities of the CMDP, such that the true transition probabilities lie in them with high probability. Formally, for any $(s,a) \in \state \times \A$, we define:
\begin{align*}
    B^{k}_{p}(s,a) = \{ \tilde{p}(.|s,a) \in \Delta_S : \forall s' \in \state, \quad   
    |\tilde{p}(s'|s,a) - \bar{p}^{k}(s'|s,a) | \leq \beta^{k}_{p}(s,a,s') \},  
\end{align*}
where the size of the confidence intervals $\beta^{k}_{p}(s,a,s')$ is built using the empirical Bernstein inequality \cite{maurer2009empirical}. For any $(s,a,s') \in \state \times \A \times \state$, it is defined as:
\begin{align*}
\beta^{k}_{p}(s,a,s') &= \sqrt{\frac{2\bar{p}^{k}(s'|s,a)(1 - \bar{p}^{k}(s'|s,a)) \ln\frac{4}{\delta'}}{\max(1,n^{k}(s,a))}} +  \frac{7 \ln \frac{4}{\delta'}}{3\max(1,n^{k}(s,a)-1)},
\end{align*}
where $\delta'$ is as defined in the algorithm and $\bar{p}^{k}(s'|s,a)(1 - \bar{p}^{k}(s'|s,a))$ is the variance associated with the empirical estimate $\bar{p}^{k}(s'|s,a)$. 
 
Given the confidence intervals $B^{k}_{p}$, the algorithm then computes a policy $\pi^{k}$ by performing optimistic planning. Given a confidence set of possible transition models, it selects an optimistic transition probability model and optimistic policy with respect to the given constrained MDP problem. This can be expressed as the following optimization problem:
\begin{align}\label{eq:optcmdp}
  (\tilde{p}^{k},\pi^{k})= \underset{\pi \in \Pi, \tilde{p} \in B^{k}_{p} }{\text{ argmin }} \quad & V_{1}^{\pi}(s_{1};c,\tilde{p})\\ \textrm{s.t.} \quad  & V_{1}^{\pi}(s_{1};d_{i},\tilde{p}) \leq l_{i} \quad \forall i \in \left[ 1: I \right].\nonumber
\end{align}
We allow time-dependent transitions, i.e., choosing different transition models at different time steps of an episode, even if the true CMDP has stationary transition probability. This does not affect the theoretical guarantees, since the true transition probability still lies in the confidence sets with high probability.

These confidence intervals differ from the ones considered in UCFH \cite{dann2015sample} which have an additional condition that the standard deviation associated with a transition model, i.e., $\sqrt{\tilde{p}(1-\tilde{p})}$ must be close to that of the empirical estimate  $\sqrt{\bar{p}(1-\bar{p})}$. We remove this condition to be able to express the optimistic planning problem \eqref{eq:optcmdp} as a linear program. However, this causes the PAC bound to have a quadratic dependence on $C$ instead of a linear dependence. 

\begin{algorithm}[H]
\caption{{\tt UC-CFH}: Upper-Confidence Constrained Fixed-Horizon episodic reinforcement learning algorithm}
\begin{algorithmic}[1]
\State \textbf{Input}: Desired tolerance $\epsilon \in (0; 1]$, failure tolerance $\delta \in (0; 1]$, fixed-horizon MDP $\M$

\State \textbf{Result}: With probability at least $1- \delta$, $\epsilon$-optimal policy \\

\State $ k := 1, \quad w_{min} = \frac{\epsilon}{4H|\state||\A|}, \quad \delta' = \frac{\delta}{2N_{\max}C};$ 
\State $N_{max} = |\state| |\A| \log_{2}\frac{|\state|H}{w_{min}};$\\
\State $ m = \frac{2304 C^{2} H^{2}}{\epsilon^2}(\log_{2} \log_{2} H)^{2} \log_{2}^{2}\frac{8H^{2}|\state|^{2}|\A|}{\epsilon}
\ln \frac{4}{\delta'};$
\State $n(s,a) = v(s,a)=n(s,a,s') = 0, $
\State $\forall s \in \state, a \in \A, s' \in Succ(s,a);$\\

\While{True}\\

\State $\bar{p}(s'|s,a) = \frac{n(s,a,s')}{\max\{1,n(s,a,s')\}},$  \State $\forall s \in \state, a \in \A, s' \in Succ(s,a)$;\\
\State $\pi^{k} = \textsc{ConstrainedExtendedLP}(\bar{p},n);$\\
\Repeat
\For{t = 0 to H-1}
\State $a_t \sim \pi^{k}_{h}(s_t);$
\State $s_{t+1} \sim p(.|s_t,a_t);$
\State $v(s_t,a_t) = v(s_t,a_t) + 1 ;$ 
\State$v(s_t,a_t,s_{t+1}) = v(s_t,a_t,s_{t+1}) + 1; $
\EndFor
\Until{there is a $(s,a) \in \state \times \A$, \\
$\text{ with } v(s,a) \geq \max\{ mw_{min},n(s,a)\} \text{ and }$\\ $n(s,a) < |\state|mH$}\\
\State $n(s,a) = n(s,a) + v(s,a);$
\State $n(s,a,s') = n(s,a,s') + v(s,a,s');$
\State $v(s,a) = v(s,a,s') = 0,$
\State $ \forall s' \in Succ(s,a);$
\State $k=k+1;$\\
\EndWhile
\end{algorithmic}
\end{algorithm}

\paragraph*{\textsc{ConstrainedExtendedLP }Algorithm.} Problem \eqref{eq:optcmdp} can be expressed as an extended LP by leveraging the state-action-state occupancy measure  $z^{\pi}(s,a,s';p)$ defined as $z^{\pi}_{h}(s,a,s';p) = p_{h}(s'|s,a)q^{\pi}_{h}(s,a;p)$ \cite{efroni2020exploration} to express the confidence intervals of the transition probabilities. The extended LP over $z$ is as follows:
\allowdisplaybreaks\begin{align*}
&  \underset{z}{\text{ min }} \quad   \sum_{h,s,a,s'}  z_{h}(s,a,s') c_{h}({s,a}),   \\
 \textrm{s.t.} \quad &z_{h}(s,a,s') \geq 0, \quad \forall (s,a,s',h) \in \state \times \A \times \state \times \left[1: H \right] \\
 &\sum_{h,s,a,s'} z_{h}(s,a,s')d_{i,h}(s,a) \leq l_{i}, \quad \forall i \in \left[ 1:I \right] \\
 &\sum_{a,s'}  z_{h}(s,a,s') = \sum_{s',a'}  z_{h}(s',a',s), \quad \forall s \in \state , \quad \forall h \in \left[2: H \right] \\
 &\sum_{a,s'}  z_{h}(s,a,s') = \mathbb{I}(s = s_1), \quad \forall s \in \state \\
  &z_{h}(s,a,s') - (\bar{p}^{k}(s'|s,a) + \beta^{k}_{p}(s,a,s'))\sum_{y}  z_{h}(s,a,y) \leq 0, \quad \forall (s,a,s',h) \in \state \times \A \times \state \times \left[1: H \right] \\
  &-z_{h}(s,a,s') + (\bar{p}^{k}(s'|s,a) - \beta^{k}_{p}(s,a,s'))\sum_{y}  z_{h}(s,a,y) \leq 0,\quad \forall (s,a,s',h) \in \state \times \A \times \state \times \left[ 1:H \right]. 
\end{align*}
The last two constraints of the above LP encode the condition that the transition probability must lie in the desired confidence interval. The desired policy $\pi^{k}$ and the chosen transition probabilities are recovered from the computed occupancy measures as:
$$ \pi^{k}_{h}(a|s) = \frac{\sum_{s'}  z_{h}(s,a,s')}{\sum_{a,s'}  z_{h}(s,a,s')} ,\quad \tilde{p}^{k}_{h}(s'|s,a) = \frac{z_{h}(s,a,s')}{\sum_{s'}  z_{h}(s,a,s')}.$$  
The above planning is referred to as  \textsc{ConstrainedExtendedLP} in the algorithm. Such an approach was also used in \cite{jin2019learning,rosenberg2019online} in the context of adversarial MDPs. The following theorem establishes the PAC guarantee for the algorithm {\tt UC-CFH}.

\begin{theorem}
For $\epsilon,\delta \in (0,1)$, with probability at least $1 - \delta$, algorithm {\tt UC-CFH} yields at most $\tilde{\mathcal{O}}\big(\frac{|\state||\A|C^{2}H^{2}}{\epsilon^{2}}\log\frac{1}{\delta}\big)$
episodes with $\epsilon$-suboptimal policies $\pi^{k}$, i.e., $V_{1}^{\pi^{k}}(s_{1},c) - V^{*} > \epsilon$ or  $V_{1}^{\pi^{k}}(s_{1},d_i) - l_i > \epsilon$, for any $i \in \left[1:I\right]$.
\end{theorem}
Thus, in the natural setting of a limited size of successor states i.e., $C \ll |S|$, the number of episodes needed by algorithm {\tt UC-CFH } to obtain an $\epsilon$-optimal policy with high probability has a linear dependence on the state and action space sizes $|S|$ and $|A|$, respectively, and quadratic dependence on the time horizon $H$.

\section{PAC Analysis}

For state-action pairs, we now introduce a notion of \emph{knownness} indicating how often the pair has been visited relative to its expected number of visits under a policy and a notion of \emph{importance} indicating the influence that the pair has on the total expected cost of a policy ~\cite{dann2015sample}. We consider a fine grained categorization of \textit{knownness} of state-action pairs similar to \cite{lattimore2012pac,dann2015sample} instead of the binary categorization \cite{brafman2002r,strehl2009reinforcement}. These are essential for the analysis of the algorithm. 

We define the weight of a state-action pair $(s, a)$ under policy $\pi^{k}$ as its expected number of visits in an episode, i.e.,
\begin{align*}
 w_{k}(s,a) &= \sum_{t=1}^{H}Pr\left[ s_t = s, a_t = a|\pi^{k},s_1\right].
\end{align*}

The \emph{importance} $\iota_{k}$ of a state-action pair $(s,a)$ with respect to policy $\pi^{k}$ is an integer defined as its relative weight with respect to $w_{min}$ on a log scale:
\begin{align*}
\iota_{k}(s,a) = \min \left\{ z_i : z_i \geq \frac{w_{k}(s,a)}{w_{min}} \right\},
\end{align*}
where $z_1 = 0, z_i = 2^{i-2}$, $\forall \ i \geq 2.$

Similarly, \emph{knownness} $\kappa_{k}$ of a state-action pair $(s,a)$ is an integer defined as:
\begin{align*}
\kappa_{k}(s,a) = \max \left\{ z_i : z_i \leq \frac{n_{k}(s,a)}{m w_{k}(s,a)} \right\},
\end{align*}
where $z_1 = 0, z_i = 2^{i-2}$, $\forall \ i \geq 2$. We then divide the $(s,a)$-pairs into categories as follows:
\begin{align*}
&\X_{k,\kappa,\iota} = \{(s,a)\in\X_{k} : \kappa_{k}(s,a) = \kappa, \iota_{k}(s,a) = \iota \}, \\&\bar{\X}_{k} = \state\times\A \backslash \X_{k},    
\end{align*}
where $\X_{k} = \{(s,a) \in \state\times\A: \iota_{k}(s,a) > 0 \} $ is the active set and $\bar{\X}_{k}$ is the inactive set, i.e., the set of state-action pairs that are unlikely to be visited under policy $\pi^{k}$. The idea is that the model estimated by the algorithm is accurate if only a small number of state-action pairs are in categories with low knownness, that is, they are important under the current policy but have not yet been sufficiently observed. 

We therefore distinguish between phases $k$ where the condition $|\X_{k,\kappa,\iota}| \leq \kappa$ for all $\kappa$ and $\iota$ holds and phases where this does not hold. This condition ensures that the number of state-action pairs in categories with low knownness are small and there are more state-action pairs in categories with higher knownness. We will further prove that the policy is $\epsilon$-optimal in episodes which satisfy this condition.

\subsection{Proof of Theorem 1}

The proof of Theorem 1 consists of the following parts: We first show in Lemma~\ref{lem:trueprob} that the true transition model is contained within the confidence sets for all phases with high probability, i.e., the true transition probability $p$ belongs to $B^{k}_{p}$ for all $k$ with probability at least $1-\frac{\delta}{2}$. Presentation of the technical lemmas used in the proof is postponed to the next subsection to improve readability.

We then use a result from \cite{dann2015sample} restated as Lemma~\ref{num_badepisodes} (with minor modification to accommodate randomized policies instead of deterministic policies) which  provides a high probability upper bound on the number of episodes for which the condition $\forall \kappa,\iota : |\X_{k,\kappa,\iota}| \leq \kappa$ is violated. Thus, the number of episodes with $|\X_{k,\kappa,\iota}| > \kappa$ for some $\kappa, \iota$ is bounded above by $6 N E_{max}$, where $N = |\state| |\A|m$ and $E_{max} = \log_{2}\frac{H}{w_{min}}\log_{2}|\state||\A|$ with probability at least $1-\frac{\delta}{2}$ (Note that the choice of $m$ in Theorem 1 satisfies the condition on $m$ in Lemma~\ref{num_badepisodes}). Thus, with high probability, i.e., at least $1-\frac{\delta}{2}$, we have $|\X_{k,\kappa,\iota}| \leq \kappa$ for all $\kappa, \iota$ for the remaining episodes. 

Thus, by union bound, for episodes beyond the first $6 N E_{max}$, we have that $|\X_{k,\kappa,\iota}| \leq \kappa$ for all $\kappa, \iota$ and $p \in B^{k}_{p}$ with probability at least $1-\delta$.

 Further, in Lemma~\ref{lem:main}, we show that in episodes with $|\X_{k,\kappa,\iota}| \leq \kappa$ for all $\kappa, \iota$, the optimistic expected total cost is $\epsilon$-close to the true expected total cost. Thus,
 $$ | V_{1}^{\pi^{k}}(s_{1},c) - \tilde{V}_{1}^{\pi^{k}}(s_{1},c)| \leq \epsilon, $$
 $$| V_{1}^{\pi^{k}}(s_{1},d_i) - \tilde{V}_{1}^{\pi^{k}}(s_{1},d_i)| \leq \epsilon, \quad \forall i \in \left[1:I\right]. $$
  We note that $\tilde{p}^{k},\pi^{k}$ were obtained by solving the following optimization problem:
\begin{align}\label{eq:opt}
  (\tilde{p}^{k},\pi^{k})= \underset{\pi \in \Pi, \tilde{p} \in B^{k}_{p} }{\text{ argmin }} \quad & V_{1}^{\pi}(s_{1};c,\tilde{p})\\ \textrm{s.t.} \quad  & V_{1}^{\pi}(s_{1};d_{i},\tilde{p}) \leq l_{i} \quad \forall i \in \left[ 1:I \right].\nonumber
\end{align}
Thus, for $p \in B^{k}_{p}$, we have,
\begin{equation*}
\begin{aligned}
 V_{1}^{\pi^{k}}(s_{1},c) - V^{*} &=   V_{1}^{\pi^{k}}(s_{1},c) - \tilde{V}_{1}^{\pi^{k}}(s_{1},c) + \tilde{V}_{1}^{\pi^{k}}(s_{1},c)- V^{*}  \\
 & \leq V_{1}^{\pi^{k}}(s_{1},c) - \tilde{V}_{1}^{\pi^{k}}(s_{1},c) \quad  (\text{By (\ref{eq:opt}), since } p \in B^{k}_{p} )\\
 &\leq \epsilon \quad \text{(By Lemma \ref{lem:main}).}
\end{aligned}    
\end{equation*}
Similarly for all $i \in \left[1:I\right]$,
\begin{equation*}
\begin{aligned}
 V_{1}^{\pi^{k}}(s_{1},d_i) - l_i &=   V_{1}^{\pi^{k}}(s_{1},d_i) - \tilde{V}_{1}^{\pi^{k}}(s_{1},d_i) +  \tilde{V}_{1}^{\pi^{k}}(s_{1},d_i)- l_i  \\
 & \leq V_{1}^{\pi^{k}}(s_{1},d_i) - \tilde{V}_{1}^{\pi^{k}}(s_{1},d_i) \quad (\text{Since $\pi^{k}$ satisfies constraints of (\ref{eq:opt})})\\
 &\leq \epsilon \quad \text{(By Lemma \ref{lem:main}).}
\end{aligned}    
\end{equation*}

Thus, putting all the above together we have that with probability at least $1-\delta$, {\tt UC-CFH} has at most $6|\state||\A|m \log_{2}\frac{H}{w_{min}}\log_{2}|\state||\A|$  $\epsilon$-suboptimal episodes.  

\subsection{Technical Lemmas}

We state the main lemmas used in the proof of Theorem 1. 
\vspace{0.5 cm}
\subsubsection{Capturing the true transition model with high probability}
We first restate the following lemma which provides an upper bound on the total number of phases in the algorithm {\tt UC-CFH} from \cite{dann2015sample}.
\vspace{0.5cm}
\begin{lemma}\label{lem:changepolicy} 
The total number of phases in the algorithm is bounded above by $N_{max} = |\state| |\A |\log_{2}\frac{|\state|H}{w_{min}}$.
\end{lemma}
\vspace{0.5cm}
The above result is used along with concentration results based on empirical Bernstein inequality \cite{maurer2009empirical} and union bounds to show that the true transition model is contained within the confidence sets for all phases with high probability.
\vspace{0.5cm}
\begin{lemma}\label{lem:trueprob}
The true transition probability is contained within the confidence intervals for all phases with high probability, i.e., $p \in B^{k}_{p}, \quad \forall k$ with probability at least $ 1 - \frac{\delta}{2}$.

\begin{proof}

Following from \cite{maurer2009empirical}, we have,

Let $\rv{Z} = (\rv{Z}_1 \ldots \rv{Z}_n)$ be independent random variables with values in $\left[ 0, 1\right]$ and let $0 < \delta < 1$. Then, with probability at least $ 1 - \delta$, we have:

\begin{align*}
    \E\left[\frac{1}{n}\sum_{i=1}^{n}\rv{Z}_{i}\right] - \frac{1}{n}\sum_{i=1}^{n}\rv{Z}_{i} \leq \sqrt{\frac{V_{n}(\rv Z) \ln\frac{2}{\delta}}{n}} + \frac{7 \ln \frac{2}{\delta}}{3(n-1)},
\end{align*}
where $V_{n}(\rv Z)$ is the sample variance,
$$ V_{n}(\rv Z) = \frac{1}{n(n-1)}\sum_{1\leq i < j \leq n}\frac{(\rv Z_i - \rv Z_j)^2}{2} .$$
By symmetry and union bound, this implies that with probability at least $1 - 2\delta$,
\begin{align*}
    | \E\left[\frac{1}{n}\sum_{i=1}^{n}\rv{Z}_{i}\right] - \frac{1}{n}\sum_{i=1}^{n}\rv{Z}_{i}| \leq \sqrt{\frac{V_{n}(\rv Z) \ln\frac{2}{\delta}}{n}} + \frac{7 \ln \frac{2}{\delta}}{3(n-1)}.
\end{align*}
That is, with probability at least $1 - \delta$, 
\begin{align}\label{eq:concentraitation}
    | \E\left[\frac{1}{n}\sum_{i=1}^{n}\rv{Z}_{i}\right] - \frac{1}{n}\sum_{i=1}^{n}\rv{Z}_{i}| \leq \sqrt{\frac{V_{n}(\rv Z) \ln\frac{4}{\delta}}{n}} + \frac{7 \ln \frac{4}{\delta}}{3(n-1)}.
\end{align}

For a single $(s,a)$ pair, $s' \in Succ(s,a)$ and phase $k$, we can consider the event that $s'$ is the next state of the MDP when choosing action $a$ in state $s$ as a Bernoulli random variable with probability $p(s'|s,a)$. Thus, by \eqref{eq:concentraitation} for $n^{k}(s,a) > 1$  (and trivially true for $n^{k}(s,a) =0,1$), with probability at least $1 - \delta'$, 
\begin{align}\label{eq:prob_ineq}
    | \bar{p}^{k}(s'|s,a) - p(s'|s,a)| &\leq \sqrt{\frac{2\bar{p}^{k}(s'|s,a)(1 - \bar{p}^{k}(s'|s,a)) \ln\frac{4}{\delta'}}{n^{k}(s,a)}} + \frac{7 \ln \frac{4}{\delta'}}{3(n^{k}(s,a)-1)} \\
    &\leq \sqrt{\frac{2\bar{p}^{k}(s'|s,a)(1 - \bar{p}^{k}(s'|s,a)) \ln\frac{4}{\delta'}}{max(1,n^{k}(s,a))}} + \frac{7 \ln \frac{4}{\delta'}}{3max(1,n^{k}(s,a)-1)},
\end{align}
 where $\bar{p}^{k}(s'|s,a)$ is the empirical estimate of the transition probability $p(s'|s,a)$ at phase $k$. ( $V_{n}(\rv Z)$ of \eqref{eq:concentraitation} simplifies to $2\bar{p}^{k}(s'|s,a)(1 - \bar{p}^{k}(s'|s,a)) $ in this case). 
 
There are at most $N_{max}$ updates or phases by Lemma~\ref{lem:changepolicy} and in each phase, a single $(s,a)$ pair with at most $C$ successor states is updated. Therefore, there are at most $N_{max}C$ such inequalities to consider. Thus, by setting $\delta' = \frac{\delta}{2CN_{max}}$ and by using union bound, the lemma is proved.
\end{proof}
\end{lemma}
\vspace{0.5 cm}
The above lemma  implies that the extended LP of the planning stage is feasible in all phases with high probability, since the true CMDP is feasible by Assumption 1.
\vspace{0.5 cm}
\subsubsection{Number of episodes which violate $|\X_{k,\kappa,\iota}| \leq \kappa, \forall \kappa,\iota $ }

We restate the following result from \cite{dann2015sample} (with minor modification to accommodate randomized policies instead of deterministic policies) which  provides a high probability upper bound on the number of episodes for which $|\X_{k,\kappa,\iota}| \leq \kappa, \forall \kappa,\iota $ is violated.

\vspace{0.5cm}
\begin{lemma}\label{num_badepisodes}
Let $E$ be the number of episodes for which there are $\kappa, \iota$ with $|\X_{k,\kappa,\iota}| > \kappa$ and and let $m \geq \frac{6H^{2}}{\epsilon}\ln\frac{2E_{max}}{\delta}$. Then,
$$\P(E \leq 6 N E_{max}) \geq 1 - \delta/2,$$ where $N = |\state| |\A|m$ and $E_{max} = \log_{2}\frac{H}{w_{min}}\log_{2}|\state||\A|$.
\end{lemma}

\vspace{0.5cm}

\subsubsection{Difference between true and optimistic total cost}

We use the following value difference lemma \cite{efroni2020exploration} to express the difference in value functions of policy $\pi$ at time step $h$ with respect to MDPs of different transition probabilities $p,\tilde{p}$, i.e., $ V_{h}^{\pi} - \tilde{V}_{h}^{\pi}$  in terms of the value functions beyond $h$, $\tilde{V}_{t}^{\pi}, t > h$ and difference in transition probabilities $(p_t - \tilde{p}_{t}), t > h $ as follows. We use shorthand $V_{h}^{\pi}(s;c), \tilde{V}_{h}^{\pi}(s;c)$ for $V_{h}^{\pi}(s;c,p), \tilde{V}_{h}^{\pi}(s;c,\tilde{p})$ respectively in the next lemma and further. Cost function $c$ is omitted when clear.

\vspace{0.5cm}

\begin{lemma}\label{lem:valdiff}
Consider MDPs $M = (\state,\A,p = \{p_{h}\}_{h=1}^{H},c = \{c_{h}\}_{h=1}^{H})$ and $\tilde{M} = (\state,\A,\tilde{p} = \{\tilde{p}_{h}\}_{h=1}^{H},c = \{c_{h}\}_{h=1}^{H}))$. Then, the difference in the values with respect to the same policy $\pi$ for any $s,h$ can be written as:
\begin{align}\label{eq:pdiff}
& V_{h}^{\pi}(s) - \tilde{V}_{h}^{\pi}(s) = 
\E \left[ \sum_{i=h}^{H}(p_{i}(\cdot|s_i,a_i)- \tilde{p}_{i}(\cdot|s_i,a_i))\tilde{V}_{h+1}^{\pi}| \pi,p,s_{h} = s \right]. 
\end{align}
\begin{proof}
The statement is trivially true for $h=H+1$ ($V_{H+1}^{\pi}(s), \tilde{V}_{H+1}^{\pi}(s) = 0 $). Let us assume it holds true for $h+1$.Then,
\allowdisplaybreaks\begin{align*}
    V_{h}^{\pi}(s) - \tilde{V}_{h}^{\pi}(s) & \\
    &= \E \left[c_{h}(s_h,a_h) + p_{h}(\cdot|s_h,a_h)V_{h+1}^{\pi}| \pi,s_h=s \right] -  \E \left[c_{h}(s_h,a_h) + \tilde{p}_{h}(\cdot|s_h,a_h)\tilde{V}_{h+1}^{\pi}| \pi,s_h=s \right] \\
    &\text{(By Bellman equation) }\\
    &= \E \left[p_{h}(\cdot|s_h,a_h)V_{h+1}^{\pi}| \pi,s_h=s \right] -  \E \left[ \tilde{p}_{h}(\cdot|s_h,a_h)\tilde{V}_{h+1}^{\pi}| \pi,s_h=s \right] \\
    &+ \E \left[p_{h}(\cdot|s_h,a_h)\tilde{V}_{h+1}^{\pi}| \pi,s_h=s \right] - \E \left[p_{h}(\cdot|s_h,a_h)\tilde{V}_{h+1}^{\pi}| \pi,s_h=s \right] \\
    &=\E \left[p_{h}(\cdot|s_h,a_h)(V_{h+1}^{\pi} -\tilde{V}_{h+1}^{\pi} )| \pi,s_h=s \right] + \E \left[ (p_{h}(\cdot|s_h,a_h)- \tilde{p}_{h}(\cdot|s_h,a_h))\tilde{V}_{h+1}^{\pi}| \pi,s_h=s \right]\\
    &=\E \left[V_{h+1}^{\pi} - \tilde{V}_{h+1}^{\pi}| \pi,p,s_h=s \right]
    + \E \left[ (p_{h}(\cdot|s_h,a_h)- \tilde{p}_{h}(\cdot|s_h,a_h))\tilde{V}_{h+1}^{\pi}| \pi,s_h=s \right]\\
    &\text{By assumption,}\\
    &= \E \left[ \left[ \sum_{i=h+1}^{H}(p_{i}(\cdot|s_i,a_i)- \tilde{p}_{i}(\cdot|s_i,a_i))\tilde{V}_{h+1}^{\pi}| \pi,p,s_{h+1} \right] |\pi,p,s_{h} = s \right] \\
    & + \E \left[ (p_{h}(\cdot|s_h,a_h)- \tilde{p}_{h}(\cdot|s_h,a_h))\tilde{V}_{h+1}^{\pi}| \pi,s_h=s \right]\\
    &=\E \left[ \sum_{i=h}^{H}(p_{i}(\cdot|s_i,a_i)- \tilde{p}_{i}(\cdot|s_i,a_i))\tilde{V}_{h+1}^{\pi}| \pi,p,s_{h} = s \right].
\end{align*}
Hence, proved by induction.
\end{proof}
\end{lemma}
\vspace{0.5cm}
We prove the following lemma which is used to upper bound the difference in transition probability $|p-\tilde{p}|$ in \eqref{eq:pdiff} in terms of $\tilde{p}$ and visitation counts $n$. The lemma is proved by viewing \eqref{eq:pineq} as a quadratic inequality in terms of $\sqrt{\bar{p}}$ and solving for $\bar{p}$. The resulting inequality is then substituted back in the original inequality to get the desired result.
\begin{lemma}\label{lem:pineq}
Let $\bar{p},\tilde{p},p \in \left[0,1 \right]$, $\delta \in (0,1)$ such that $p,\tilde{p} \in CI$ where, 
\begin{align}\label{eq:pineq}
CI := &\{ p' \in \left[0,1 \right]: |p' - \bar{p} | \leq \sqrt{\frac{2\bar{p}(1 - \bar{p}) \ln\frac{4}{\delta}}{\max(1,n)}} + \frac{7 \ln \frac{4}{\delta}}{3\max(1,n-1)} \}.
\end{align}

Then, $|\tilde{p} - p| \leq 2\sqrt{2}\sqrt{\frac{\tilde{p}\ln\frac{4}{\delta}}{max(1,n-1)}} + 5(\frac{\ln\frac{4}{\delta}}{max(1,n-1)})^{\frac{3}{4}} + \frac{21 \ln \frac{4}{\delta}}{max(1,n-1)}$.

\begin{proof}
The lemma is trivially true for $n=0,1$. For $n>1$ and $p \in CI$, we have,
\allowdisplaybreaks\begin{align}\label{eq:comp_square}
    |p - \bar{p} | &\leq \sqrt{\frac{2\bar{p}(1 - \bar{p}) \ln\frac{4}{\delta}}{n}} + \frac{7 \ln \frac{4}{\delta}}{3(n-1)} \nonumber \\
    &\leq 2\sqrt{\frac{\bar{p} \ln\frac{4}{\delta}}{2(n-1)}} + \frac{7 \ln \frac{4}{\delta}}{3(n-1)}.
\end{align}
For $\tilde{p} \in CI$, we similarly have,
\allowdisplaybreaks\begin{align}\label{eq:comp_square1}
    |\tilde{p} - \bar{p} | &\leq 2\sqrt{\frac{\bar{p} \ln\frac{4}{\delta}}{2(n-1)}} + \frac{7 \ln \frac{4}{\delta}}{3(n-1)}.
\end{align}
By simplifying \eqref{eq:comp_square}, we get,
\allowdisplaybreaks\begin{align}\label{eq:ptrueineq}\nonumber
    \bar{p} \leq p + 2\sqrt{\frac{\bar{p} \ln\frac{4}{\delta}}{2(n-1)}} +  \frac{7 \ln \frac{4}{\delta}}{3(n-1)} \\ \nonumber
   \implies (\sqrt{\bar{p}})^{2} \leq (\sqrt{p})^{2} + 2 \sqrt{\bar{p}}\sqrt{\frac{ \ln\frac{4}{\delta}}{2(n-1)}}+  \frac{7 \ln \frac{4}{\delta}}{3(n-1)} \\ \nonumber
   \implies (\sqrt{\bar{p}})^{2} - 2 \sqrt{\bar{p}}\sqrt{\frac{ \ln\frac{4}{\delta}}{2(n-1)}} + \frac{\ln \frac{4}{\delta}}{2(n-1)} \leq (\sqrt{p})^{2} + \frac{17 \ln \frac{4}{\delta}}{6(n-1)}\\ \nonumber
   \implies (\sqrt{\bar{p}} - \sqrt{\frac{ \ln\frac{4}{\delta}}{2(n-1)}} )^{2} \leq (\sqrt{p})^{2}  + (\sqrt{\frac{17 \ln \frac{4}{\delta}}{6(n-1)}})^{2}\\ \nonumber
   \text{ Since for }a,b \geq 0,  a^2 + b^2 \leq (a+b)^2, \\ \nonumber
   \implies |\sqrt{\bar{p}} - \sqrt{\frac{ \ln\frac{4}{\delta}}{2(n-1)}} | \leq \sqrt{p} + \sqrt{\frac{17 \ln \frac{4}{\delta}}{6(n-1)}} \\ 
   \implies \sqrt{\bar{p}} \leq \sqrt{p} + (\frac{1}{\sqrt{2}} + \sqrt{\frac{17}{6}})\sqrt{\frac{\ln \frac{4}{\delta}}{(n-1)}}.
\end{align}
Substituting \eqref{eq:ptrueineq} in \eqref{eq:comp_square}, we get,
\allowdisplaybreaks\begin{align*}
    |p - \bar{p} | &\leq \sqrt{\frac{2\ln\frac{4}{\delta}}{(n-1)}}(\sqrt{p} + (\frac{1}{\sqrt{2}} + \sqrt{\frac{17}{6}})\sqrt{\frac{\ln \frac{4}{\delta}}{(n-1)}}) + \frac{7 \ln \frac{4}{\delta}}{3(n-1)}\\
    &\leq \sqrt{\frac{2p\ln\frac{4}{\delta}}{(n-1)}} + (1+\sqrt{\frac{17}{3}})\frac{ \ln \frac{4}{\delta}}{(n-1)} + \frac{7 \ln \frac{4}{\delta}}{3(n-1)}\\
    &= \sqrt{\frac{2p\ln\frac{4}{\delta}}{(n-1)}} + (\frac{10}{3}+\sqrt{\frac{17}{3}})\frac{ \ln \frac{4}{\delta}}{(n-1)}.
\end{align*}
Similarly, substituting \eqref{eq:ptrueineq} in \eqref{eq:comp_square1}, we get,
\allowdisplaybreaks\begin{align*}
    |\tilde{p} - \bar{p} | &\leq \sqrt{\frac{2p\ln\frac{4}{\delta}}{(n-1)}} + (\frac{10}{3}+\sqrt{\frac{17}{3}})\frac{ \ln \frac{4}{\delta}}{(n-1)}
\end{align*}
Thus,
\allowdisplaybreaks\begin{align}\label{eq:tilde}
    |\tilde{p} - p| &\leq   |\tilde{p} - \bar{p}| + |\bar{p} - p| \nonumber \\
    &\leq 2\Big( \sqrt{\frac{2p\ln\frac{4}{\delta}}{(n-1)}} +  (\frac{10}{3}+\sqrt{\frac{17}{3}})\frac{ \ln \frac{4}{\delta}}{(n-1)}\Big)   \\
    &= 2\sqrt{2}\sqrt{\frac{\ln\frac{4}{\delta}}{(n-1)}}(\sqrt{p - \tilde{p}+ \tilde{p}}) + (\frac{20}{3}+2\sqrt{\frac{17}{3}})\frac{ \ln \frac{4}{\delta}}{(n-1)}\nonumber \\
&\leq 2\sqrt{2}\sqrt{\frac{\ln\frac{4}{\delta}}{(n-1)}}(\sqrt{|p - \tilde{p}|+ \tilde{p}}) + (\frac{20}{3}+2\sqrt{\frac{17}{3}})\frac{ \ln \frac{4}{\delta}}{(n-1)}\nonumber \\
\nonumber
\end{align}
\begin{align*}
\text{ Since for }a,b &\geq 0,  \sqrt{a + b} \leq \sqrt{a
}+\sqrt{b}, \text{ and using \eqref{eq:tilde}},\\
&\leq 2\sqrt{2}\sqrt{\frac{\tilde{p}\ln\frac{4}{\delta}}{(n-1)}} + 4\sqrt{\frac{\ln\frac{4}{\delta}}{(n-1)}}\sqrt{\sqrt{\frac{2p\ln\frac{4}{\delta}}{(n-1)}} + (\frac{10}{3}+\sqrt{\frac{17}{3}})\frac{ \ln \frac{4}{\delta}}{(n-1)}} \\
&+ (\frac{20}{3}+2\sqrt{\frac{17}{3}})\frac{ \ln \frac{4}{\delta}}{(n-1)}\\
&\leq 2\sqrt{2}\sqrt{\frac{\tilde{p}\ln\frac{4}{\delta}}{(n-1)}} + 4\sqrt{\frac{\ln\frac{4}{\delta}}{(n-1)}}\Big( (\frac{2p\ln\frac{4}{\delta}}{(n-1)})^{\frac{1}{4}} + \sqrt{\frac{10}{3}+\sqrt{\frac{17}{3}}} \sqrt{\frac{ \ln \frac{4}{\delta}}{(n-1)}}\Big) \\
&+ (\frac{20}{3}+2\sqrt{\frac{17}{3}})\frac{ \ln \frac{4}{\delta}}{(n-1)}\\
&\leq 2\sqrt{2}\sqrt{\frac{\tilde{p}\ln\frac{4}{\delta}}{(n-1)}} + 4\sqrt{\frac{\ln\frac{4}{\delta}}{(n-1)}}\Big( (\frac{2\ln\frac{4}{\delta}}{(n-1)})^{\frac{1}{4}} + \sqrt{\frac{10}{3}+\sqrt{\frac{17}{3}}} \sqrt{\frac{ \ln \frac{4}{\delta}}{(n-1)}}\Big) \\
&+ (\frac{20}{3}+2\sqrt{\frac{17}{3}})\frac{ \ln \frac{4}{\delta}}{(n-1)}\\
&\leq 2\sqrt{2}\sqrt{\frac{\tilde{p}\ln\frac{4}{\delta}}{(n-1)}} + 5(\frac{\ln\frac{4}{\delta}}{(n-1)})^{\frac{3}{4}} + \frac{21 \ln \frac{4}{\delta}}{(n-1)}.
\end{align*}
\end{proof}
\end{lemma}

For each phase $k$, the true transition probability $p$ belongs to the confidence set $B^{k}_{p}$ with high probability and the optimistic transition model $\tilde{p}^{k}$ is chosen from the confidence set and thus $p$ and $\tilde{p}^{k}$ belong to $CI$ for suitable $\delta$, by definition of $B^{k}_{p}$.
Therefore, by Lemma \ref{lem:pineq}, $|p - \tilde{p}^{k}|$ can be upper bounded in terms of $\tilde{p}^{k}$ and $n$ as described above.

The following lemma upper bounds the summand in \eqref{eq:pdiff},  $(p-\tilde{p}_{h})(\cdot|s,a)\tilde{V}_{h+1}$ which is the difference of the expected value of successor states in MDPs with true transition probability $p$ and optimistic transition probability model $\tilde{p}$. 

\begin{lemma}\label{lem:varineq}
Let, $$ |p(s'|s,a) - \tilde{p}_{h}(s'|s,a)| \leq c_{1}(s,a) + c_{2}(s,a)\sqrt{\tilde{p}_{h}(s'|s,a)},$$ for all $s,s' \in \state$ and $a \in \A $. Then, for any policy $\pi$,
$$|(p-\tilde{p}_{h})(\cdot|s,a)\tilde{V}_{h+1}| \leq c_{1}(s,a) |Succ(s,a)| \|\tilde{V}_{h+1} \|_{\infty} + c_{2}(s,a)\sqrt{|Succ(s,a)|}\tilde{\sigma}_{h}(s,a), $$ for any $(s,a) \in \state \times \A$, where $\tilde{\sigma}_{h}^{2}$ is the local variance function defined as:
$$\tilde{\sigma}_{h}^{2}(s,a) = \E \left[ (\tilde{V}_{h+1}(s_{h+1}) - \E(\tilde{V}_{h+1}(s_{h+1})|s_{h}=s,\tilde{p},\pi))^{2}|s_h =s, a_h = a, \tilde{p}\right]. $$
\begin{proof}
Let $\hat{V}(s') = \E(\tilde{V}_{h+1}(s_{h+1})|s_{h}=s,\tilde{p},\pi) $ be a constant function. Then,
\allowdisplaybreaks\begin{align*}
    |(p-\tilde{p}_{h})(\cdot|s,a)\tilde{V}_{h+1}| &= |(p-\tilde{p}_{h})(\cdot|s,a)(\tilde{V}_{h+1} + \hat{V} - \hat{V})|\\
    &= |(p-\tilde{p}_{h})(\cdot|s,a)(\tilde{V}_{h+1} - \hat{V})| \\
    & (\text{ as } (p-\tilde{p}_{h})(\cdot|s,a)\hat{V} = 0 \text{ as } \hat{V} \text{ is a constant function.})\\
    &\leq \sum_{s' \in Succ(s,a)}|p(s'|s,a) - \tilde{p}(s'|s,a)||\tilde{V}_{h+1}(s') - \hat{V}(s')| \text{ (By triangular inequality)} \\
    &\leq \sum_{s' \in Succ(s,a)} (c_{1}(s,a) + c_{2}(s,a)\sqrt{\tilde{p}_{h}(s'|s,a)})|\tilde{V}_{h+1}(s') - \hat{V}(s')| \\
    &\leq c_{1}(s,a) |Succ(s,a)| \|\tilde{V}_{h+1} \|_{\infty} \text{ (as value is non negative)} \\
    &+ c_{2}(s,a)\sum_{s' \in Succ(s,a)}\sqrt{\tilde{p}_{h}(s'|s,a)(\tilde{V}_{h+1}(s') - \hat{V}(s'))^{2}}\\
    &\leq c_{1}(s,a) |Succ(s,a)| \|\tilde{V}_{h+1} \|_{\infty}+ c_{2}(s,a)\sqrt{|Succ(s,a)|\sum_{s' \in Succ(s,a)}\tilde{p}_{h}(s'|s,a)(\tilde{V}_{h+1}(s') - \hat{V}(s'))^{2}}\\
    &\text{ (By Cauchy-Schwartz inequality) }\\
    &= c_{1}(s,a) |Succ(s,a)| \|\tilde{V}_{h+1} \|_{\infty} + c_{2}(s,a)\sqrt{|Succ(s,a)|}\tilde{\sigma}_{h}(s,a).
\end{align*}
Hence, proved.
\end{proof}
\end{lemma}
\vspace{0.5cm}
We then consider a sequence of MDPs $\M^{(d)}$ which have the same transition probability as that of the true MDP, i.e., $p$ but different cost functions $c^{(d)}$. A similar sequence of MDPs $\tilde{\M}^{(d)}$ which have the same transition probability $\tilde{p}$ is also considered. 

In both sequences, for $d=0$, the cost function is the same as that of the original cost function, i.e., $c^{(0)}_{h},\tilde{c}^{(0)}_{h} = c_{h}, 1 \leq h \leq H$. The following cost functions are then defined recursively as $c^{(2d+2)}_{h}(s,a),\tilde{c}^{(2d+2)}_{h}(s,a) = \tilde{\sigma}_{h}^{(d),2}(s) $ where $\tilde{\sigma}_{h}^{(d),2}$ is the local variance of the value function under policy $\pi$ with respect to the costs $c^{(d)}$ and defined as:
\begin{align*}
\tilde{\sigma}_{h}^{(d),2}(s) = \E \left[ (\tilde{V}_{h+1}^{(d)}(s_{h+1}) - \E(\tilde{V}_{h+1}^{(d)}(s_{h+1}) |s_{h}=s,\tilde{p},\pi))^{2}|s_h =s, \pi, \tilde{p}\right].     
\end{align*}
Note that $c^{(d)}_{h}(s,a) \in \left[0, H^{d}\right]$. We use the notation $V^{(d)}$ and $\tilde{V}^{(d)}$ for value functions of $\M^{(d)}$ and $\tilde{\M}^{(d)}$ respectively.

We also use the following lemma \cite{dann2015sample} to bound $\sum_{i=1}^{H}\E\left[\tilde{\sigma}^{2}_{i}(s_{i})|s_{h}=s,\tilde{p},\pi \right]$ in Lemma \ref{lem:recursive} by $O(H^2)$ instead of the trivial $O(H^3)$.

\begin{lemma}\label{lem:valuevariance}
The variance of the value function defined as $\mathcal{V}_{h}^{\pi}(s) = \E\left[(\sum_{i=h}^{H} c_{i}(s_i,a_i) - V_{i}^{\pi}(s_i))^{2}|s_{h} = s,\pi \right]$ satisfies a Bellman equation $\mathcal{V}_{h}(s) = \E\left[\mathcal{V}_{h}(s_{h+1})|s_{h}=s,\pi \right]  + \sigma^{2}_{h}(s)$ which gives $\mathcal{V}_{h}(s) = \sum_{i=h}^{H} \E\left[\sigma^{2}_{i}(s_{i})|s_{h}=s,\pi \right] $. Since $0 \leq \mathcal{V}_{1} \leq H^{2}c_{max}^{2}$, we have $0\leq \sum_{i=1}^{H}\E\left[\sigma^{2}_{i}(s_{i})|s_{h}=s,\pi \right] \leq  H^{2}c_{max}^{2}$ for all $s \in \state$.
\end{lemma} 
\vspace{0.5cm}

If $p,\tilde{p} \in B^{k}_{p}$, the condition of Lemma \ref{lem:varineq} holds true by Lemma \ref{lem:pineq} for suitable constants. Then, by utilizing Lemmas \ref{lem:valdiff}, \ref{lem:varineq} and \ref{lem:valuevariance}, we have the following recursive relation relating $| V_{1}^{(d)}(s_{1}) - \tilde{V}_{1}^{(d)}(s_{1})|$ with $| V_{1}^{(d+1)}(s_{1}) - \tilde{V}_{1}^{(d+1)}(s_{1})|$ when the condition $|\X_{\kappa,\iota}| \leq \kappa$ for all $(\kappa,\iota)$ holds true. The analysis follows by splitting the state action pairs by importance, i.e., $(s,a) \in \X$ and $(s,a) \not\in \X$ and using the definitions of weight $w$, knownness $\kappa$ and importance $\iota$. 

\begin{lemma}\label{lem:recursive}
Let $p,\tilde{p} \in B^{k}_{p}$. If $|\X_{\kappa,\iota}| \leq \kappa$ for all $(\kappa,\iota)$. Then,
$$| V_{1}^{(d)}(s_{1}) - \tilde{V}_{1}^{(d)}(s_{1})| := \Delta_{d} \leq \hat{A}_{d} + \hat{B}_{d}^{1} + \hat{B}_{d}^{2} + \text{ min}\{\hat{C}_{d},\hat{C}'_{d} + \hat{C}'' \sqrt{\Delta_{2d+2} }\}   $$ where, 
$$ \hat{A}_{d} = \frac{\epsilon H^{d}}{4 }, \quad \hat{B}_{d}^{1} = 42 CH^{d+1}\big( \frac{|\mathcal{K}\times\mathcal{I}|\ln \frac{4}{\delta'}}{m}\big), \quad \hat{B}_{d}^{2} = 10 CH^{d+5/4}\big( \frac{|\mathcal{K}\times\mathcal{I}|\ln \frac{4}{\delta'}}{m}\big)^{3/4},  $$
$$  \hat{C}'_{d} = \sqrt{\frac{16C|\mathcal{K}\times\mathcal{I}|}{m}\ln\frac{4}{\delta'} H^{2d+2}},\quad \hat{C}_{d} = \hat{C}'_{d} \sqrt{H}, \quad \hat{C}'' = \sqrt{\frac{16C|\mathcal{K}\times\mathcal{I}|}{m}\ln\frac{4}{\delta'}}.  $$
\begin{proof}
\allowdisplaybreaks\begin{align*}
\Delta_{d} &= | V_{1}^{(d)}(s_{1}) - \tilde{V}_{1}^{(d)}(s_{1})| \\
&= \big{|}\E \left[ \sum_{h=1}^{H}(p-\tilde{p}_{h})(\cdot|s_h,a_h)\tilde{V}_{h+1}^{(d)}| \pi,p,s_1 \right]\big{|} \text{ (By Lemma \ref{lem:valdiff})}\\
&\leq \sum_{h=1}^{H} \E \left[ \big{|}(p-\tilde{p}_{h})(\cdot|s_h,a_h)\tilde{V}_{h+1}^{(d)}\big{|}| \pi,p,s_1 \right] \text{ (By triangular inequality and Jensen's inequality)}\\
&= \sum_{h=1}^{H} \E \left[ \sum_{s,a} \mathbb{I}\{s_h = s, a_h =a \} \big{|}(p-\tilde{p}_{h})(\cdot|s_h,a_h)\tilde{V}_{h+1}^{(d)}\big{|}| \pi,p,s_1 \right]\\
&= \sum_{s,a} \sum_{h=1}^{H} \E \left[  \mathbb{I}\{s_h = s, a_h =a \} \big{|}(p-\tilde{p}_{h})(\cdot|s,a)\tilde{V}_{h+1}^{(d)}\big{|}| \pi,p,s_1 \right]\\
&= \sum_{s,a} \sum_{h=1}^{H} \big{|}(p-\tilde{p}_{h})(\cdot|s,a)\tilde{V}_{h+1}^{(d)}\big{|} \E \left[  \mathbb{I}\{s_h = s, a_h =a \} | \pi,p,s_1 \right] \\
&= \sum_{s,a \not\in \X} \sum_{h=1}^{H} \big{|}(p-\tilde{p}_{h})(\cdot|s,a)\tilde{V}_{h+1}^{(d)}\big{|} \E \left[  \mathbb{I}\{s_h = s, a_h =a \} | \pi,p,s_1 \right] \\
&+ \sum_{s,a \in \X} \sum_{h=1}^{H} \big{|}(p-\tilde{p}_{h})(\cdot|s,a)\tilde{V}_{h+1}^{(d)}\big{|} \E \left[  \mathbb{I}\{s_h = s, a_h =a \} | \pi,p,s_1 \right] \\
&\leq \sum_{s,a \not\in \X} \sum_{h=1}^{H} \|\tilde{V}_{h+1}^{(d)}\|_{\infty} \E \left[  \mathbb{I}\{s_h = s, a_h =a \} | \pi,p,s_1 \right] \\
&+\sum_{s,a \in \X} \sum_{h=1}^{H} \big{(}c_{1}(s,a) |Succ(s,a)| \|\tilde{V}_{h+1}^{(d)} \|_{\infty} + c_{2}(s,a)\sqrt{|Succ(s,a)|}\tilde{\sigma}_{h}^{(d)}(s,a)\big{)} \E \left[  \mathbb{I}\{s_h = s, a_h =a \} | \pi,p,s_1 \right] \\
&\text{ (By Lemma \ref{lem:varineq}, with $c_{1}(s,a),c_{2}(s,a)$ obtained from Lemma \ref{lem:pineq})} \\
& \leq \sum_{s,a \not\in \X} \sum_{h=1}^{H} H^{d+1} \E \left[  \mathbb{I}\{s_h = s, a_h =a \} | \pi,p,s_1 \right] \\
&+\sum_{s,a \in \X} \sum_{h=1}^{H} \big{(}c_{1}(s,a) |Succ(s,a)| H^{d+1} + c_{2}(s,a)\sqrt{|Succ(s,a)|}\tilde{\sigma}_{h}^{(d)}(s,a)\big{)} \E \left[  \mathbb{I}\{s_h = s, a_h =a \} | \pi,p,s_1 \right] \\\\
&= \sum_{s,a \not\in \X} H^{d+1} w(s,a) + \sum_{s,a \in \X}c_{1}(s,a) |Succ(s,a)| H^{d+1} w(s,a) \\
&+ \sum_{s,a \in \X}c_{2}(s,a)\sqrt{|Succ(s,a)|}\sum_{h=1}^{H} \tilde{\sigma}_{h}^{(d)}(s,a)\E \left[  \mathbb{I}\{s_h = s, a_h =a \} | \pi,p,s_1 \right] \\
& \text{(By definition of $w(s,a)$)} \\
&\leq \underbrace{\sum_{s,a \not\in \X} H^{d+1} w(s,a)}_{A(s_{1})} + \underbrace{\sum_{s,a \in \X}c_{1}(s,a) C H^{d+1} w(s,a) }_{B(s_{1})}\\
&+ \underbrace{\sum_{s,a \in \X}c_{2}(s,a)\sqrt{C}\sum_{h=1}^{H} \tilde{\sigma}_{h}^{(d)}(s,a)\E \left[  \mathbb{I}\{s_h = s, a_h =a \} | \pi,p,s_1 \right].}_{C(s_1)}\\
& \text{(Since $C$ is an upper bound of $|Succ(s,a)|, \forall s,a.$)}
\end{align*}
By assumption that $p,\tilde{p} \in B^{k}_{p} $, we have from Lemma \ref{lem:pineq}, 
$$ c_{2}(s,a) = 2\sqrt{2}\sqrt{\frac{\ln\frac{4}{\delta'}}{max(1,n(s,a)-1)}}, \quad c_{1}(s,a) = 5(\frac{\ln\frac{4}{\delta'}}{max(1,n(s,a)-1)})^{\frac{3}{4}} + \frac{21 \ln \frac{4}{\delta'}}{max(1,n(s,a)-1)}.$$ 
Using these, we now simply $A(s_1),B(s_1),C(s_1)$.
\begin{align*}
    A(s_1) &= \sum_{s,a \not\in \X} H^{d+1} w(s,a) \\
        & (\text{ as } w(s,a) \leq w_{min} \quad \forall s,a \not\in \X  )\\
        &\leq w_{min}H^{d+1}|\state||\A|\\
        &= \frac{\epsilon H^{d+1}|\state||\A|}{4 H |\state||\A|} = \frac{\epsilon H^{d}}{4 } \\
        &=\hat{A}_{d}.\\ \\
    B(s_1) &= \sum_{s,a \in \X}c_{1}(s,a) C H^{d+1} w(s,a) \\
    &= C H^{d+1}\left[ \sum_{s,a \in \X}5(\frac{\ln\frac{4}{\delta'}}{max(1,n(s,a)-1)})^{\frac{3}{4}} w(s,a) + \sum_{s,a \in \X}\frac{21 \ln \frac{4}{\delta'}w(s,a)}{max(1,n(s,a)-1)}   \right].
\end{align*}
Now,
\begin{align*}
    \sum_{s,a \in \X} \frac{w(s,a)}{max(1,n(s,a)-1)} &\leq \sum_{\kappa,\iota}\sum_{s,a \in \X_{\kappa,\iota}}
    \frac{w(s,a)}{max(1,n(s,a)-1)} \\
    &=\sum_{\kappa,\iota}\sum_{s,a \in \X_{\kappa,\iota}}
    \frac{w(s,a)}{n(s,a)}\frac{n(s,a)}{max(1,n(s,a)-1)}\\
    \text{ For } s,a \in \X_{\kappa,\iota}, \text{ we have } n(s,a) &\geq mw(s,a)\kappa. \text{ Thus, } \frac{w(s,a)}{n(s,a)} \leq \frac{1}{\kappa m}.\\ 
    \text{ Further, since }  |\X_{\kappa,\iota}| \leq \kappa, \text{ we have for} &\text{ all relevant } (s,a) \text{ pairs, } n(s,a) > 0\\ 
    &\leq \sum_{\kappa,\iota}\sum_{s,a \in \X_{\kappa,\iota}} \frac{1}{\kappa m}\cdot 2 \\
    &\leq \sum_{\kappa,\iota} \frac{2}{m} \text{ (as } |\X_{\kappa,\iota}| \leq \kappa) \\
    &=\frac{2|\mathcal{K} \times \mathcal{I}|}{m}.
\end{align*}
Continuing,
\begin{align*}
    \sum_{s,a \in \X} \frac{w(s,a)}{(max(1,n(s,a)-1))^{3/4}} &\leq   \sum_{\kappa,\iota}\sum_{s,a \in \X_{\kappa,\iota}} \frac{w(s,a)}{(max(1,n(s,a)-1))^{3/4}} \\
    &= \sum_{\kappa,\iota}\sum_{s,a \in \X_{\kappa,\iota}} \frac{w(s,a)}{(n(s,a))^{3/4}}\frac{(n(s,a))^{3/4}}{(max(1,n(s,a)-1))^{3/4}} \\
    &\leq 2 \sum_{\kappa,\iota}\sum_{s,a \in \X_{\kappa,\iota}} \frac{w(s,a)}{(n(s,a))^{3/4}} \text{ (as before) } \\
    &= 2 \sum_{\kappa,\iota}\sum_{s,a \in \X_{\kappa,\iota}} (w(s,a))^{1/4}(\frac{w(s,a)}{n(s,a)})^{3/4}\\
    &\leq 2 \sum_{\kappa,\iota}\sum_{s,a \in \X_{\kappa,\iota}}(w(s,a))^{1/4} (\frac{1}{\kappa m})^{3/4} \text{ (as before) } \\
    &= 2 \sum_{\kappa,\iota}\sum_{s,a \in \X_{\kappa,\iota}} (\frac{1}{\kappa m})^{1/2} (\frac{w(s,a)}{\kappa m})^{1/4}\\
    &\leq 2\sum_{\kappa,\iota}\sqrt{\frac{|\X_{\kappa,\iota}|}{\kappa m}\sum_{s,a \in \X_{\kappa,\iota}}(\frac{w(s,a)}{\kappa m})^{1/2}} \text{ (By Cauchy-Schwartz inequality) }\\
    &\leq \frac{2}{m^{1/2}}\sum_{\kappa,\iota}\sqrt{\sum_{s,a \in \X_{\kappa,\iota}}(\frac{w(s,a)}{\kappa m})^{1/2}} \text{ (as } |\X_{\kappa,\iota}| \leq \kappa)\\
    &\leq \frac{2 |\mathcal{K}\times \mathcal{I}|^{1/2}}{m^{1/2}}\sqrt{\sum_{\kappa,\iota}\sum_{s,a \in \X_{\kappa,\iota}}(\frac{w(s,a)}{\kappa m})^{1/2}}\text{ (By Cauchy-Schwartz inequality) }\\
    &= \frac{2 |\mathcal{K}\times \mathcal{I}|^{1/2}}{m^{3/4}}\sqrt{\sum_{\kappa,\iota}\sum_{s,a \in \X_{\kappa,\iota}}(\frac{w(s,a)}{\kappa})^{1/2}}\\
    &\leq \frac{2 |\mathcal{K}\times \mathcal{I}|^{1/2}}{m^{3/4}}\sqrt{\sum_{\kappa,\iota}\sqrt{\frac{|\X_{\kappa,\iota}|}{\kappa}\sum_{s,a \in \X_{\kappa,\iota} }w(s,a)}}\text{ (By Cauchy-Schwartz inequality) }\\
    &\leq \frac{2 |\mathcal{K}\times \mathcal{I}|^{1/2}}{m^{3/4}}\sqrt{\sum_{\kappa,\iota}\sqrt{\sum_{s,a \in \X_{\kappa,\iota} }w(s,a)}} \text{ (as } |\X_{\kappa,\iota}| \leq \kappa)\\
    &\leq \frac{2 |\mathcal{K}\times \mathcal{I}|^{1/2}}{m^{3/4}}\sqrt{\sqrt{|\mathcal{K} \times \mathcal{I} |\sum_{\kappa,\iota}\sum_{s,a \in \X_{\kappa,\iota}}w(s,a)}} \text{ (By Cauchy-Schwartz inequality) }\\
    &\leq \frac{2 |\mathcal{K}\times \mathcal{I}|^{3/4}}{m^{3/4}} H^{1/4}.
\end{align*}
Thus, putting together the above two, we get that,
$$B(s_1) \leq  \underbrace{42 CH^{d+1}\big( \frac{|\mathcal{K}\times\mathcal{I}|\ln \frac{4}{\delta'}}{m}\big)}_{\hat{B}_{d}^{1}} + \underbrace{10 CH^{d+5/4}\big( \frac{|\mathcal{K}\times\mathcal{I}|\ln \frac{4}{\delta'}}{m}\big)^{3/4}.}_{\hat{B}_{d}^{2}} $$
Now,
\begin{align*}
    C(s_1) &= \sum_{s,a \in \X}c_{2}(s,a)\sqrt{C}\sum_{h=1}^{H} \tilde{\sigma}_{h}^{(d)}(s,a)\E \left[  \mathbb{I}\{s_h = s, a_h =a \} | \pi,p,s_1 \right] \\
    &\leq \sqrt{C} \sum_{s,a \in \X} c_{2}(s,a) \sqrt{\sum_{h=1}^{H}\E \left[  \mathbb{I}\{s_h = s, a_h =a \} | \pi,p,s_1 \right]} \\& \times\sqrt{\sum_{h=1}^{H}\tilde{\sigma}_{h}^{(d),2}(s,a)\E \left[  \mathbb{I}\{s_h = s, a_h =a \} | \pi,p,s_1 \right]}\text{ (By Cauchy-Schwartz inequality) }\\
    &=\sqrt{C} \sum_{s,a \in \X} \sqrt{\frac{8w(s,a)}{max(1,n(s,a)-1)}\ln\frac{4}{\delta'}\sum_{h=1}^{H}\tilde{\sigma}_{h}^{(d),2}(s,a)\E \left[  \mathbb{I}\{s_h = s, a_h =a \} | \pi,p,s_1 \right]}\\
    &(\text{ By definition of } w(s,a) \text{ and } c_{2}(s,a) \text{ from Lemma \ref{lem:pineq}} )\\
    &\leq \sqrt{C} \sum_{\kappa,\iota}\sum_{s,a \in \X_{\kappa,\iota}} \sqrt{\frac{8w(s,a)}{max(1,n(s,a)-1)}\ln\frac{4}{\delta'}\sum_{h=1}^{H}\tilde{\sigma}_{h}^{(d),2}(s,a)\E \left[  \mathbb{I}\{s_h = s, a_h =a \} | \pi,p,s_1 \right]}\\
    &\leq \sqrt{C} \sum_{\kappa,\iota} \sqrt{|\X_{\kappa,\iota}|\sum_{s,a \in \X_{\kappa,\iota}}\frac{8w(s,a)}{max(1,n(s,a)-1)}\ln\frac{4}{\delta'}\sum_{h=1}^{H}\tilde{\sigma}_{h}^{(d),2}(s,a)\E \left[  \mathbb{I}\{s_h = s, a_h =a \} | \pi,p,s_1 \right]}\\
    &\text{ (By Cauchy-Schwartz inequality) }\\
    &\leq \sqrt{C} \sum_{\kappa,\iota} \sqrt{\sum_{s,a \in \X_{\kappa,\iota}}\frac{16}{m}\ln\frac{4}{\delta'}\sum_{h=1}^{H}\tilde{\sigma}_{h}^{(d),2}(s,a)\E \left[  \mathbb{I}\{s_h = s, a_h =a \} | \pi,p,s_1 \right]} \text{ (as before )}\\
    &\leq\sqrt{\frac{16C|\mathcal{K}\times\mathcal{I}|}{m}\ln\frac{4}{\delta'}\sum_{s,a \in \X}\sum_{h=1}^{H}\tilde{\sigma}_{h}^{(d),2}(s,a)\E \left[  \mathbb{I}\{s_h = s, a_h =a \} | \pi,p,s_1 \right]}\text{ (By Cauchy-Schwartz inequality) }\\
    &\leq\sqrt{\frac{16C|\mathcal{K}\times\mathcal{I}|}{m}\ln\frac{4}{\delta'}\sum_{s,a \in \state \times \A}\sum_{h=1}^{H}\tilde{\sigma}_{h}^{(d),2}(s,a)\E \left[  \mathbb{I}\{s_h = s, a_h =a \} | \pi,p,s_1 \right]}\\
    &=\sqrt{\frac{16C|\mathcal{K}\times\mathcal{I}|}{m}\ln\frac{4}{\delta'}\sum_{s,a \in \state \times \A}\sum_{h=1}^{H}\tilde{\sigma}_{h}^{(d),2}(s,a)\P  (s_h = s, a_h =a  | \pi,p,s_1 )}\\
    &=\sqrt{\frac{16C|\mathcal{K}\times\mathcal{I}|}{m}\ln\frac{4}{\delta'}\sum_{h=1}^{H}\sum_{s,a \in \state \times \A}\tilde{\sigma}_{h}^{(d),2}(s,a)\P  (s_h = s, a_h =a  | \pi,p,s_1 )}\\
    &=\sqrt{\frac{16C|\mathcal{K}\times\mathcal{I}|}{m}\ln\frac{4}{\delta'}\sum_{h=1}^{H}\E \left[\tilde{\sigma}_{h}^{(d),2}(s_h)| \pi,p,s_1 \right]}\\
    & \text{ Since } \|\tilde{\sigma}_{h}^{(d),2} \|_{\infty} \leq H^{2d+2}, \text{ we have, }\\
    C(s_1) &\leq \sqrt{\frac{16C|\mathcal{K}\times\mathcal{I}|H^{2d+3}}{m}\ln\frac{4}{\delta'}} = \hat{C}_{d}.
\end{align*}
Else,
\vspace{0.5cm}
\allowdisplaybreaks\begin{align*}
    C(s_1) &\leq \sqrt{\frac{16C|\mathcal{K}\times\mathcal{I}|}{m}\ln\frac{4}{\delta'}\sum_{h=1}^{H}\E \left[\tilde{\sigma}_{h}^{(d),2}(s_h)| \pi,p,s_1 \right]}\\
    &= \sqrt{\frac{16C|\mathcal{K}\times\mathcal{I}|}{m}\ln\frac{4}{\delta'}\sum_{h=1}^{H}(\E \left[\tilde{\sigma}_{h}^{(d),2}(s_h)| \pi,p,s_1 \right]-\E \left[\tilde{\sigma}_{h}^{(d),2}(s_h)| \pi,\tilde{p},s_1 \right]+\E \left[\tilde{\sigma}_{h}^{(d),2}(s_h)| \pi,\tilde{p},s_1 \right])}\\
    &\text{ By Lemma \ref{lem:valuevariance}, we have that } \sum_{h=1}^{H} \E \left[\tilde{\sigma}_{h}^{(d),2}(s_h)| \pi,\tilde{p},s_1 \right] = \tilde{\mathcal{V}}^{(d)}_{1}(s_1) \leq H^{2d+2}. \text{ Using this,}\\
    &\leq \sqrt{\frac{16C|\mathcal{K}\times\mathcal{I}|}{m}\ln\frac{4}{\delta'}\big( H^{2d+2} + V^{(2d+2)}_{1}(s_1) - \tilde{V}^{(2d+2)}_{1}(s_1) \big)}\\
    &\leq \sqrt{\frac{16C|\mathcal{K}\times\mathcal{I}|}{m}\ln\frac{4}{\delta'}\big( H^{2d+2} + \Delta_{2d+2} \big)}\\
    &\leq \underbrace{\sqrt{\frac{16C|\mathcal{K}\times\mathcal{I}|}{m}\ln\frac{4}{\delta'} H^{2d+2}}}_{\hat{C}'_{d}} +  \underbrace{\sqrt{\frac{16C|\mathcal{K}\times\mathcal{I}|}{m}\ln\frac{4}{\delta'}}}_{\hat{C}''}\sqrt{\Delta_{2d+2} } .\\ &(\text{ Since for }a,b \geq 0,  \sqrt{a + b} \leq \sqrt{a}+\sqrt{b})
\end{align*}
Putting all the above together, the lemma is proved.
\end{proof}
\end{lemma}
\vspace{0.5cm}
This recurrence relation is simplified to show in Lemma~\ref{lem:main} that in phases with $|\X_{k,\kappa,\iota}| \leq \kappa$ for all $\kappa,\iota$, the optimistic total expected cost $\tilde{V}_{1}^{\pi^{k}}(s_{1})$  is close to that of the true one, $V_{1}^{\pi^{k}}(s_{1})$. This lemma plays an important role in the final theorem to show that the policy obtained after sufficiently large number of episode is $\epsilon$-optimal with respect to the objective and constraints.
\vspace{0.5cm}
\begin{lemma}\label{lem:main}
Let $p,\tilde{p} \in B^{k}_{p}$. If $|\X_{k,\kappa,\iota}| \leq \kappa$ for all $\kappa,\iota$ and $0 < \epsilon \leq 1 $ and 
$$ m \geq  \frac{2304 C^{2} H^{2}}{\epsilon^2}(\log_{2} log_{2} H)^{2} \log_{2}^{2}\frac{8H^{2}|\state|^{2}|\A|}{\epsilon}
\ln \frac{4}{\delta'},$$ then $| V_{1}^{\pi^{k}}(s_{1}) - \tilde{V}_{1}^{\pi^{k}}(s_{1})| \leq \epsilon$.\\
\begin{proof}
From Lemma \ref{lem:recursive}, we have the following recursive relation,
$$ \Delta_{d} \leq \hat{A}_{d} + \hat{B}_{d}^{1} + \hat{B}_{d}^{2} + \hat{C}'_{d} + \hat{C}'' \sqrt{\Delta_{2d+2} },$$
which is the of the form $ \Delta_{d} \leq Y_d + Z\sqrt{\Delta_{2d+2} } $. Expanding this recursive expression up to level $\gamma = \lceil \frac{\ln H}{2 \ln 2} \rceil$ with repeated use of $ \sqrt{a + b} \leq \sqrt{a}+\sqrt{b}, \quad \forall a,b \geq 0  $, we get,
\begin{align*}
    \Delta_{0} &\leq Y_0 + Z\sqrt{\Delta_2}\\
    &\leq Y_0 + Z\sqrt{Y_2 + Z\sqrt{\Delta_6}}\\
    &\leq Y_0 + Z\sqrt{Y_2} + Z^{3/2}\Delta_{6}^{1/4}\\
    & \leq \vdots \\
    & \leq \sum_{d \in D\backslash\{ \gamma\}} Z^{\frac{2d}{2+d}}Y_{d}^{\frac{2}{2+d}} + Z^{\frac{2\gamma}{2 + \gamma}}\Delta_{\gamma}^{\frac{2}{2+\gamma}},
\end{align*}
where $D = \{0,2,6,14,\ldots,\gamma \}$. Further, for $m \geq 110.25 C |\mathcal{K} \times \mathcal{I}| \ln \frac{4}{\delta'}$, we have $\hat{C}'_{d} \geq \hat{B}_{d}^{1} $. Similarly, for \\$m \geq 40 C^{2} H |\mathcal{K} \times \mathcal{I}| \ln \frac{4}{\delta'}$, we have $\hat{C}'_{d} \geq \hat{B}_{d}^{2} $. Thus, for $ m \geq 110.25 C^{2} H |\mathcal{K} \times \mathcal{I}| \ln \frac{4}{\delta'} $, we can set, 
$$ Y_{d} = \hat{A}_{d} + 3\hat{C}'_{d},  $$
$$Z = \hat{C}''.$$
Further, since $\hat{C}_{d} = \hat{C}'_{d} \sqrt{H} \geq \hat{C}'_{d}$, for the large enough m, we have,
\begin{align*}
    \Delta_{\gamma} &\leq \hat{A}_{\gamma} + 2\hat{C}'_{\gamma} + \text{ min}\{ \hat{C}_{\gamma}, \hat{C}'_{\gamma} + \hat{C}'' \sqrt{\Delta_{2\gamma+2}} \}  \\
    &\leq \hat{A}_{\gamma} +3\hat{C}_{\gamma}.
\end{align*}
Let $ m_{1} = \frac{16C|\mathcal{K} \times \mathcal{I} | H^{2} }{m \epsilon^{2}}\ln\frac{4}{\delta'}$ and thus get,
\allowdisplaybreaks\begin{align*}
    Z &= \hat{C}'' = \frac{\sqrt{m_{1}}\epsilon}{H}, \\
    Y_{d} &= \hat{A}_{d} + 3\hat{C}'_{d} = (\frac{1}{4} + 3\sqrt{m_1})H^{d}\epsilon, \\
    \Delta_{\gamma} &\leq \hat{A}_{\gamma} +3\hat{C}_{\gamma} = (\frac{1}{4} + 3\sqrt{m_{1}H})H^{d}\epsilon.
\end{align*}
Thus,
\begin{align*}
    (Z^{2d}Y_{d}^{2})^{(2+d)^{-1}} &= (m_{1}^{d}\epsilon^{2d+2}(\frac{1}{4}+3\sqrt{m_{1}})^{2})^{(2+d)^{-1}} \\
    &= \epsilon(m_{1}^{d}\epsilon^{d}(\frac{1}{4}+3\sqrt{m_{1}})^{2})^{(2+d)^{-1}}, \\
    (Z^{2\gamma}\Delta_{\gamma}^{2})^{(2+\gamma)^{-1}} &=(m_{1}^{\gamma}\epsilon^{2\gamma+2}(\frac{1}{4}+3\sqrt{m_{1}H})^{2})^{(2+\gamma)^{-1}} \\
&= \epsilon(m_{1}^{\gamma}\epsilon^{\gamma}(\frac{1}{4}+3\sqrt{m_{1}H})^{2})^{(2+\gamma)^{-1}}. \\
\end{align*}
Putting the above together, we have,
\begin{align*}
    \frac{\Delta_{0}}{\epsilon} &\leq \sum_{d \in D\backslash \{ \gamma\}} \left[ (\epsilon m_{1})^{\frac{d}{d+2}}(\frac{1}{4}+3\sqrt{m_{1}})^{\frac{2}{2+d}})\right] + (\epsilon m_{1})^{\frac{\gamma}{\gamma+2}}(\frac{1}{4}+3\sqrt{m_{1}H})^{\frac{2}{2+\gamma}})\\
    &=\frac{1}{4} + 3\sqrt{m_{1}} + \sum_{d \in D\backslash \{0, \gamma\}} \left[ (\epsilon m_{1})^{\frac{d}{d+2}}(\frac{1}{4}+3\sqrt{m_{1}})^{\frac{2}{2+d}})\right] \\
    &+ (\epsilon m_{1})^{\frac{\gamma}{\gamma+2}}(\frac{1}{4}+3\sqrt{m_{1}H})^{\frac{2}{2+\gamma}})\\
    & \leq \frac{1}{4} + 3\sqrt{m_{1}} + \sum_{d \in D\backslash \{0, \gamma\}}  (\epsilon m_{1})^{\frac{d}{d+2}}\left[(\frac{1}{4})^{\frac{2}{2+d}}+(3\sqrt{m_{1}})^{\frac{2}{2+d}})\right] \\
    &+ (\epsilon m_{1})^{\frac{\gamma}{\gamma+2}}\left[(\frac{1}{4})^{\frac{2}{2+\gamma}}+(3\sqrt{m_{1}H})^{\frac{2}{2+\gamma}})\right]\\
    & (\text{ Since for }a,b \geq 0,\quad 0 < \phi < 1, \quad   (a + b)^{\phi} \leq {a}^{\phi}+{b}^{\phi})
\end{align*}
And now using,
$$ \frac{1}{2+\gamma} \leq \frac{1}{2 + \frac{\ln H}{2\ln 2}} = \frac{2 \ln 2}{4\ln2 + \ln H } \leq 2 \log_{H}2,$$
we bound $H^{\frac{1}{2 + \gamma}} \leq 4$. Thus,
\begin{align*}
    \frac{\Delta_{0}}{\epsilon} &\leq \frac{1}{4} + 3\sqrt{m_{1}} + \sum_{d \in D\backslash \{0, \gamma\}}  (\epsilon m_{1})^{\frac{d}{d+2}}\left[(\frac{1}{4})^{\frac{2}{2+d}}+(3\sqrt{m_{1}})^{\frac{2}{2+d}})\right] \\
    &+ (\epsilon m_{1})^{\frac{\gamma}{\gamma+2}}\left[(\frac{1}{4})^{\frac{2}{2+\gamma}}+4(3\sqrt{m_{1}})^{\frac{2}{2+\gamma}})\right]\\
    &\leq \frac{1}{4} + 3\sqrt{m_{1}} + \sum_{d \in D\backslash \{0 \}}  (\epsilon m_{1})^{\frac{d}{d+2}}\left[(\frac{1}{4})^{\frac{2}{2+d}}+4(3\sqrt{m_{1}})^{\frac{2}{2+d}})\right] \\
    &\text{(By change of parameters,)} \\
    &\leq \frac{1}{4} + 3\sqrt{m_{1}} + \sum_{i=1}^{\log_{2}\gamma}  (\epsilon m_{1})^{1-2^{-i}}\left[(\frac{1}{4})^{2^{-i}}+4(3\sqrt{m_{1}})^{2^{-i}})\right] \\
    &\leq \frac{1}{4} + 3\sqrt{m_{1}} + \sum_{i=1}^{\log_{2}\gamma}  ( m_{1})^{1-2^{-i}}\left[(\frac{1}{4})^{2^{-i}}+4(3\sqrt{m_{1}})^{2^{-i}})\right] \text{ (as } \epsilon \leq 1)\\
    & = \frac{1}{4} + 3\sqrt{m_{1}} + \sum_{i=1}^{\log_{2}\gamma}  ( m_{1})^{1-2^{-i}}\left[(4)^{-2^{-i}}+4(3\sqrt{m_{1}})^{2^{-i}})\right] \\
    &= \frac{1}{4} + 3\sqrt{m_{1}} + \frac{1}{4}\sum_{i=1}^{\log_{2}\gamma} (4m_{1})^{1-2^{-i}} + 4 \sum_{i=1}^{\log_{2}\gamma} ( m_{1})^{1-2^{-i}} ( 9m_{1})^{2^{-i-1}}\\
    &=  \frac{1}{4} + 3\sqrt{m_{1}} + \frac{1}{4}\sum_{i=1}^{\log_{2}\gamma} (4m_{1})^{1-2^{-i}} + 36 \sum_{i=1}^{\log_{2}\gamma}(\frac{m_{1}}{9})^{1-2^{-i-1}}.\\
\end{align*}
By requiring $ m_{1} \leq \frac{1}{4}$, we have $ 4m_{1} < 1 $ and $\frac{m_{1}}{9} < 1$. Further for $i \geq 1$, $ 1-2^{-i} \geq \frac{1}{2} $ and $1-2^{-i-1} \geq \frac{3}{4}$. Thus, for  $ m_{1} \leq \frac{1}{9}$,
\begin{align*}
    \frac{\Delta_{0}}{\epsilon} &\leq  \frac{1}{4} + 3\sqrt{m_{1}} + \frac{1}{4}\sqrt{4m_{1}}\log_{2}\gamma + 36(\frac{m_{1}}{9})^{3/4}\log_{2}\gamma \\
    &= \frac{1}{4} + 3\sqrt{m_{1}} + \sqrt{\frac{m_{1}}{4}}\log_{2}\gamma + 36(\frac{m_{1}}{9})^{3/4}\log_{2}\gamma. \\
\end{align*}
By requiring that $m_{1} \leq 1/144 $, $m_{1} \leq (2\log_{2}\gamma)^{-2}$ and $m_{1} \leq \frac{1}{41}(\log_{2}\gamma)^{-4/3}$,  we get $\Delta_{0} \leq \epsilon $. Thus taking \\$m_{1} \leq \frac{1}{144}(\log_{2}\gamma)^{-2} \leq \frac{1}{144}(\log_{2} log_{2} H)^{-2} $ (as $ \log_{2}\gamma = \log_{2}(\lceil \frac{\log_{2}H}{2}\rceil \leq \log_{2} log_{2} H)$) meets all our previous requirements on $m_{1}$. 
Now using the definition of $m_{1}$, we get that this implies,
$$ m \geq \frac{2304 C |\mathcal{K} \times \mathcal{I}| H^{2}}{\epsilon^2} (\log_{2} log_{2} H)^{2} \ln \frac{4}{\delta'} .$$ But, we had an earlier requirement that $m \geq 110.25 C^{2} H |\mathcal{K} \times \mathcal{I}| \ln \frac{4}{\delta'}$. Thus, we choose the stronger condition on $m$ i.e.,
$$  m \geq \frac{2304 C^{2} |\mathcal{K} \times \mathcal{I}| H^{2}}{\epsilon^2}(\log_{2} log_{2} H)^{2} \ln \frac{4}{\delta'}. $$

By construction, $\iota(s,a) \leq \frac{2H}{w_{min}} = \frac{8H^{2}|\state||\A|}{\epsilon}$. Also, $\kappa(s,a) \le \frac{|\state|mH}{mw_{min}} = \frac{4|\state|^{2}|\A|H^{2}}{\epsilon}$. 

Thus, $|\mathcal{K} \times \mathcal{I}| \leq  \log_{2}\frac{8H^{2}|\state||\A|}{\epsilon} \log_{2} \frac{4|\state|^{2}|\A|H^{2}}{\epsilon} \leq \log_{2}^{2}\frac{8H^{2}|\state|^{2}|\A|}{\epsilon}  $. Substituting this in $m$ gives us the desired result.
\end{proof}
\end{lemma}
\section{Conclusions}

Motivated by safe reinforcement learning for autonomous systems in unknown environments, we addressed the problem of finding approximately optimal policies for finite-horizon MDPs with constraints and unknown transition probability. We introduced the {\tt UC-CFH} algorithm that is based on the optimism-in-the-face-of-uncertainty principle and offered, to the best of our knowledge, the first result in terms of provable PAC guarantees for both performance and constraint violations.
Our PAC bound exhibits quadratic dependence on the horizon length. In the future, we plan to consider other types of constraints, e.g., chance or risk constraints, and extensions to the infinite-horizon setting.

\newpage
\bibliographystyle{IEEEtran}
\bibliography{refs}
\end{document}